\documentclass[sigconf]{acmart}

\usepackage{array}
\newcommand{\etal}{\textit{et al}. }

\newcommand{\eg}{\textit{e}.\textit{g}. }
\usepackage{balance}
\usepackage{siunitx} % for \ang{}
\usepackage{amsfonts} % For \mathbb
\usepackage{amsmath} % For \begin{align}
\usepackage{multirow} % for multirow in tables

\AtBeginDocument{%
  \providecommand\BibTeX{{%
    \normalfont B\kern-0.5em{\scshape i\kern-0.25em b}\kern-0.8em\TeX}}}

\setcopyright{acmcopyright}
\copyrightyear{2021}
\acmYear{2021}
\acmDOI{10.1145/3411763.3451554}

\acmConference[CHI '21]{CHI '21: The ACM CHI Conference on Human Factors in Computing Systems}{May 08--13, 2021}{Yokohama, Japan}
\acmBooktitle{CHI '21: The ACM CHI Conference on Human Factors in Computing Systems,
  May 08--13, 2021, Yokohama, Japan}
\acmPrice{}
\acmISBN{978-1-4503-8095-9/21/05}

\acmSubmissionID{1140}

\begin{document}

\title{Real-time Gesture Animation Generation from Speech for Virtual Human Interaction}

\author{Manuel Rebol}
\email{mrebol@american.edu}
%\email{webmaster@marysville-ohio.com}
\affiliation{%
  \institution{American University, Graz University Of Technology}
  %\streetaddress{P.O. Box 1212}
  \city{Graz}
  \country{Austria}
}

\author{Christian Gütl}
\email{c.guetl@tugraz.at}
\affiliation{%
  \institution{Graz University Of Technology}
  %\streetaddress{P.O. Box 1212}
  \city{Graz}
  \country{Austria}
}

\author{Krzysztof Pietroszek}
\email{pietrosz@american.edu}
\affiliation{%
  \institution{American University}
  \city{Washington}
  \country{USA}
}

\begin{abstract}
We propose a real-time system for synthesizing gestures directly from speech. Our data-driven approach is based on Generative Adversarial Neural Networks to model the speech-gesture relationship. We utilize the large amount of speaker video data available online to train our 3D gesture model. Our model generates speaker-specific gestures by taking consecutive audio input chunks of two seconds in length.  We animate the predicted gestures on a virtual avatar. We achieve a delay below three seconds between the time of audio input and gesture animation. 
Code and videos are available at \small \url{https://github.com/mrebol/Gestures-From-Speech}
\end{abstract}






    \begin{teaserfigure}
    \centering
    %\vspace{-0.5cm}
        %\includegraphics[width=.5\textwidth]{fig/gruber_full_nobg.png}%
      %\includegraphics[width=.5\textwidth]{fig/oliver_full_nobg.png}
      \includegraphics[width=0.9\textwidth]{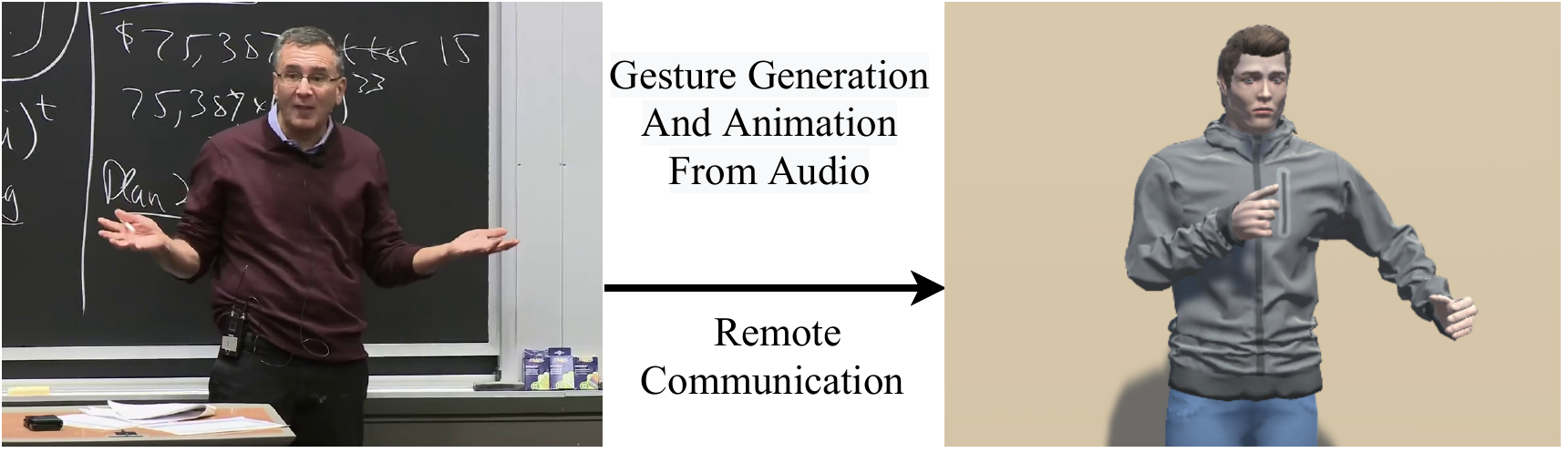}
      \caption{Real-time gesture generation for professor Jonathan G. On the left, we show the original pose in a video. Our system captures the audio and predicts gestures as shown in the right image. Our system can be used in remote communication scenarios where no video capture is possible.}
      %\Description{,,,,}
      \label{fig:teaser}
    \end{teaserfigure}

\maketitle

\section{Introduction}
% Intro and applications
Equipping digital representations of humans with non-verbal communication skills finds many applications in human-computer systems, from telehealth to entertainment to education, because non-verbal communication is an essential component of communication process. 

In our previous work, we tackled the problem of unrealistic human-agent communication by introducing a generative model that synthesizes natural-looking gestures \cite{rebol2021}. In this work, we expand the proposed model to generate gestures from speech in real time. Our approach utilizes the large amount of video data available to train a model to learn the relationship between speech and gesture.
In contrast to our previous work work \cite{rebol2021} and recent related work \cite{speech2gesture2019}, we extract and predict 3D gestures in real time, with relatively small latency. 

Our models are speaker-specific because body language is specific to every individual. We train a model individually for every speaker for that we generated a dataset. Our model also captures the individual gesturing style of speakers from different genres. We animated the individual gesture predictions on a virtual avatar to provide a natural human-like representation to the observer. 

Our real-time gesture generation system can be used in remote communication applications. Whenever there exists no possibility to video capture the speaker, our system can be used to animate a virtual human instead of capturing the speaker. By having both audio and a visual representation, a more vivid communication experience is achieved. We support to move the communication into Virtual Reality because we animate the generated gestures in a 3D environment. Our system can also be used in low-bandwidth remote communication where sending audio is possible, but video is not supported. In this scenario, the remote communication partner would be able to watch at a virtual representation of the speaker.   

%To extract the gesture from our large video dataset we use state-of-the-art human pose estimation algorithms %\cite{openpose1, videopose3d2019, hand3d2017}. We adopt them such that they estimate 3D human gestures. We encode %the speech and pass them together with the gestures to our Generative Adversarial Network (GAN) %\cite{goodfellow2014generative} model. After training the parameters of our GAN model, we animate the generated %gestures in virtual reality. %on virtual humans.
%We evaluate the quality of our animations by conducting a user study in which we compare the generated gestures %against the original gestures and against gestures from uncorrelated speech. 

\section{Our Approach}

% Structural Approach
Our approach is divided into three core components. 
%We illustrate the structural overview in \autoref{fig:gesture-pipeline}. 
Our input component provides the GAN with the gestures. We estimate the 2D human pose in each frame of input video sequences. Once the 2D human pose is estimated, it is projected into 3D and handed over to the GAN. 
The main component contains the gesture generation GAN. Our GAN takes two inputs during training, the raw audio of the speech and the gestures in human pose format. We extract the raw audio from our video dataset. 
The third component animates the generated gestures on a character in virtual reality.

\subsection{Extracting Gestures from Video}
% We need a lot of data -> 2d videos and body pose extraction
We extract the training data for our GAN model from a readily-available large source of data: 2D videos of people performing lectures and speeches. Recent advancements in 3D human pose estimation from 2D video allow us to process many online videos of lectures and speeches and extract 3D body poses and gestures from them.

% Human Pose Format% Valid frames
We encode the gestures from video in the 3D human pose format. 
We take precautions to ensure high quality of data, both before and after extracting the body language from the video. We eliminate those fragments of the videos where the hands and upper body of the speaker is not entirely visible, \eg because it is being partially occluded by an object. 

We approach the gesture extraction in two steps. In the first step, we extract the 2D Human Pose from the raw video using the OpenPose framework \cite{openpose1, openpose2}. In the second step, we project the 2D pose into 3D space for the large body parts and hands separately. For the 3D body pose estimation we use the model implemented by Pavllo \etal \cite{videopose3d2019}. For the 3D hand pose estimation we implement a model similar to Zimmerman \etal \cite{hand3d2017}. As a result of the process, we created a large dataset of motion captures that correspond to the gestures used by humans when speaking.

\subsection{Generating Gestures from Speech}
% GAN intro
We implement the Generative Adversarial Neural Network (GAN) \cite{goodfellow2014generative} framework which allows us to model the multimodal task of predicting gestures from speech. The GAN framework consists of the gesture generator $G$ and the motion discriminator $D$. 

\paragraph{Gesture Generator} The objective of the gesture generator is twofold. First, the generator is trained to predict gestures close to ground truth gestures extracted from video input. Second, the main objective of the generator inside the GAN framework is to fool the discriminator. In our case, the discriminator is fooled by predicted gestures with realistic motion. 
We implement a UNet \cite{unet} architecture to generate gestures encoded in the human pose format from speech. Inside the UNet, the skip connections forward low-level prosodic features extracted from the input audio. These features are necessary to predict smaller beat gestures.  %and lip motion accurately. 
The bottleneck extracts high-level features that contain information about long input sequences. This is useful to predict the posture of the speaker. 

The objective of the generator is enforced by a regression loss on the prediction given pseudo ground truth gestures. The loss function for the generator is defined as
\begin{align} \label{eqn:gen}
\mathcal{L}_\text{Gen}(G) \ = \ &\mathbb{E}_{(\mathbf{s},\mathbf{p})}[ ||\mathbf{p} - G(\mathbf{s})||_1 ] \ + \\
&\lambda_\text{bone} \ \mathbb{E}_{(\mathbf{s})}[ ||B(G(\mathbf{\mathbf{s}}_t)) - B(G(\mathbf{s}_{t-1}))||_1 ] ,
\end{align}
where vector $\mathbf{s}$ refers to the input speech and vector $\mathbf{p}$ refers to the pseudo ground truth body keypoints. The function $B$ computes the bone length which is computed by the euclidean distance between pairs of keypoints at consecutive time steps $t$ and $t-1$. Therefore, the second term in \autoref{eqn:gen} ensures that the bone length stays constant over time. 
The first term ensures that the predicted output matches the ground truth gestures extracted from the video. The hyperparameter $\lambda_\text{bone} \in (0,1)$ is used to weight the importance of constant bone length in the prediction. 

\paragraph{Motion Discriminator} 
% why we need the motion discriminator
Our discriminator ensures that the motion of the generated gestures is similar to the motion extracted from video to avoid regression towards the mean gesture. We compute the motion between consecutive time steps by subtracting the keypoint positions:
\begin{equation} 
\label{eqn:motion}
M(\mathbf{v}) = \mathbf{v}_{t} - \mathbf{v}_{t-1} \ .
\end{equation}
The discriminator receives either a real or a fake gesture sequences as input. 
The fake gesture sequence is predicted by the generator, whereas the real gesture sequence is directly obtained from the input video. 
The discriminator is trained to distinguish real and generated gestures. 
The complete GAN loss function including the discriminator $D$ is defined as 
\begin{equation} 
\label{eqn:gan}
\mathcal{L}_\text{GAN}(G, D) = \mathbb{E}_{(\mathbf{p})}[\log D(M(\mathbf{p})) ]+ \mathbb{E}_{(\mathbf{s})}[\log(1 - D(M(G(\mathbf{s}))))] \ .
\end{equation}
The objective of the discriminator is to maximize this function. Consequently, the term loss is only true with respect to the generator. The discriminator learns to output $D(\cdot) \rightarrow 1$ if input motion is real and $D(\cdot) \rightarrow 0$ if the input motion is generated.

\paragraph{GAN Objective}
We train the parameters of our model by combining the loss functions shown in \autoref{eqn:gen} and \autoref{eqn:gan}. 
The final objective function is defined as
\begin{equation} \label{eqn:Whole}
\min_G \max_D \mathcal{L}_\text{GAN}(G, D) + \mathcal{L}_\text{Gen}(G)	.
\end{equation}
The generator $G$ has the objective to minimize this function whereas the discriminator $D$ aims to maximize $\mathcal{L}_\text{GAN}$. 

\begin{figure*}[h]
    \centering
    \includegraphics[width=\textwidth]{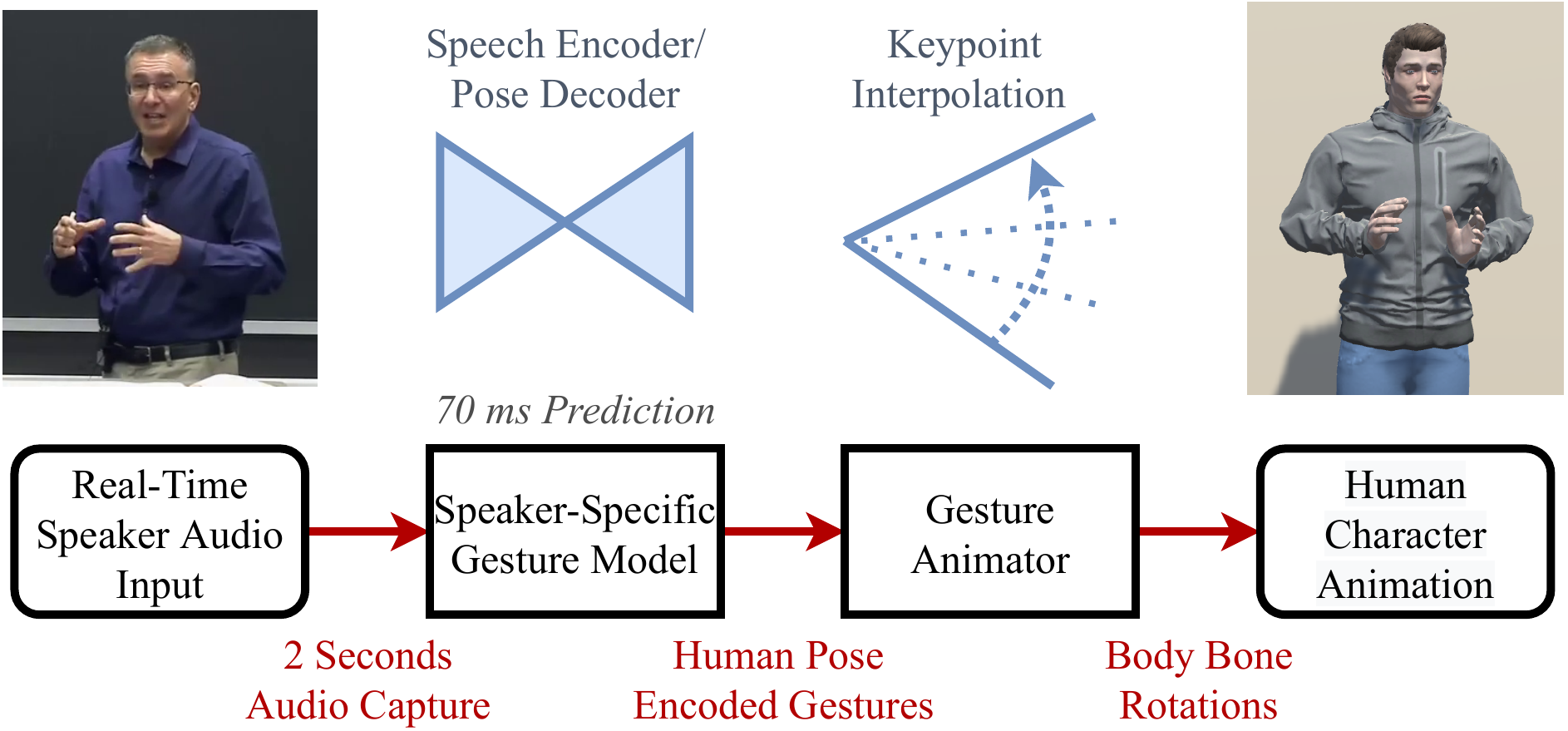}
    \caption{System diagram. Our real-time gesture prediction system takes chunks of two seconds speaker audio as an input and animates the predicted gestures on a virtual human. In this process, a speaker-specific gesture model generates gestures in the human pose format. The gesture animator translates the encoded gestures into a 3D animation.  }
    \label{fig:system}
\end{figure*}

\subsection{Animating Gestures in Virtual Reality} 
Our Gesture GAN model produces 3D Human Pose sequences which we animate on virtual humans using rotation angles between bones and inverse kinematic computation. By connecting the keypoints predicted by our Gesture GAN, we create a skeletal representation of the human pose. We use this skeletal representation of the gestures to animate an avatar in virtual reality.

One challenge when transferring the predicted 3D Human Pose into a 3D animation is the missing information about anatomical details and ambiguities. Since we animate the gestures on a virtual human of different size and shape compared to the original speaker, we omit the information about the bone length of the generated skeleton. Instead, we only consider the rotation between the bones. To tackle the problem of missing rotation information, we use inverse kinematics on the human pose input.

Besides recovering missing rotation information, we also tackle implausible motion predictions. Although our Gesture GAN is trained to predict motion which is similar to real human motion, there exists no constraint which particularly enforces anatomically plausible motion. Hence, the predicted motion in some cases appears artificial, especially when animated on a virtual human avatar. This problem is very distracting for the viewer of the animation because anatomically implausible motion is quickly noticed. To overcome this issue, we introduce motion constraints on arms and fingers to filter out implausible motion. In addition to the motion constraints, we apply motion smoothing. Both optimizations combined produce an appealing animation.  

\subsection{Training the GAN} 

For the purpose of training our GAN, we generate a gesture dataset of over one hundred hours extracted from videos of four speakers from two domains. Specifically, we generated 72 hours of motion capture data for television show hosts Oliver and Ellen as well as 64 hours of motion capture data for professor Jonathan G. and Shelly K. We encode the gestures in the efficient human pose format. By using this format, we ignore irrelevant information such as background and the shape of different body parts of the speaker.

We evaluate the predictions of our Gesture GAN on our validation set using the Percent of Correct Keypoints (PCK) \cite{pck2013} metric with proximity radius $\alpha = 0.2$. The quantitative results are 31.0 PCK for speaker Ellen, 60.3 PCK for speaker Oliver, 40.6 PCK for speaker Shelly K. and 23.6 PCK for speaker Jonathan G. We observe that Oliver and Shelly K. who are in sitting position and therefore show less upper body movement achieve a higher PCK. In contrast, the PCK is lower for speakers which are standing (Ellen) and walking around (Jonathan G.).

\subsection{System Design}
Our real-time system consists of two main components, the speaker-specific gesture prediction model and the gesture animator. We show an overview of the system in \autoref{fig:system}. The gesture prediction model contains a trained UNet model with a speech encoder and a human pose decoder. The output of the predictor are gestures in the form human pose keypoints. The human pose keypoints are taken as an input by the other component, the gesture animator. The animator reads keypoints and transforms them into animations for a virtual character. To achieve that, the keypoint positions which are located on joints are transferred into skeletal bone rotations. Incomplete information about rotation angles are estimated by using inverse kinematic computations. The motion between keypoint positions is interpolated using linear interpolation.  

We take speech sequences of two seconds as input for our real-time system. The gesture prediction model reads the two second audio input to predict gestures. The two second input provides context for the model to prediction beat gestures. Our model needs about 70 milliseconds to generate the corresponding output gestures from two second input. In our experiments we used a mid-range desktop computer setup with an Intel 8th generation i7 processor and a Nvidia Geforce GTX 1070 GPU.  The generated gestures are passed to the animator which reads a two seconds of human pose data at the rate 15 poses per second. Besides interpolation between poses from within the two second sequences, the animator also interpolates between the last pose of the previous sequence and the first pose of the current sequence. The interpolation between sequences is computed more smoothly because the pose deviation between different sequences is larger. The total delay of our system is about 2.2 seconds which results from the two second sequence length, the prediction time, and the animation interpolation.

\subsection{Future Work}
Our gesture synthesizer is able to generate gestures for a given speaker with perceptual quality that could not be determined to be significantly different than the ground truth. However, because the training depends on the sound wave of the speech, the synthesis is speaker-specific and cannot generate good quality gestures for an arbitrary speaker. To address a wider spectrum of the applications, the predictor should also be able to synthesize gesture animations given a speech stream of an arbitrary speaker. One way to address this issue is to automatically transcribe the speech from the video into text using speech recognition algorithms and preserving the resulting text temporal alignment with the motion data. The resulting $speech$ vector in our $<speech, gesture>$ training pair will now consists of text, rather than an audio sequence. This approach would significantly reduce the dimensionality of the input vector thus reducing the optimal size of the training data required.

\section{Conclusion}
We proposed a method of synthesizing gestures on avatars in real-time given input speech. Our model that translates speech to gestures is speaker-specific and it learns by observing speakers utilizing the large amount of video data available. We achieve a delay of less than three seconds when applying the gesture prediction in a real-time communication setting. This delay mainly comes from the fact that our model needs temporal context and is not able to predict gestures instantly. As a result, our system can be used most efficiently in one-way communication applications.

\section{Acknowledgments}
This work was supported in part by the Marshall Plan Foundation and the National Science Foundation.

\bibliographystyle{ACM-Reference-Format}
\bibliography{template}

\end{document}